\documentclass[conference]{IEEEtran}
\usepackage{cite}
\usepackage{amsmath,amssymb,amsfonts}
\usepackage{algorithmic}
\usepackage{graphicx}
\usepackage{textcomp}
\usepackage{xcolor}

\usepackage{adjustbox}
\usepackage[caption=false,font=footnotesize]{subfig}
\usepackage{booktabs}
\usepackage{hyperref}

\newtheorem{definition}{Definition}

\def\BibTeX{{\rm B\kern-.05em{\sc i\kern-.025em b}\kern-.08em
    T\kern-.1667em\lower.7ex\hbox{E}\kern-.125emX}}
\begin{document}


\title{WaveGNN: Integrating Graph Neural Networks and Transformers for Decay-Aware Classification of Irregular Clinical Time-Series}

\author{\IEEEauthorblockN{Arash Hajisafi, Maria Despoina Siampou, Bita Azarijoo, Zhen Xiong, Cyrus Shahabi}
\IEEEauthorblockA{\textit{Dept. of Computer Science, University of Southern California, Los Angeles, CA, USA} \\
\{hajisafi, siampou, azarijoo, xiongzhe, shahabi\}@usc.edu}
}


\maketitle

\begin{abstract}
Clinical time series are often irregularly sampled, with varying sensor frequencies, missing observations, and misaligned timestamps. Prior approaches typically address these irregularities by interpolating data into regular sequences, thereby introducing bias, or by generating inconsistent and uninterpretable relationships across sensor measurements, complicating the accurate learning of both intra-series and inter-series dependencies. We introduce WaveGNN, a model that operates directly on irregular multivariate time series without interpolation or conversion to a regular representation. WaveGNN combines a decay-aware Transformer to capture intra-series dynamics with a sample-specific graph neural network that models both short-term and long-term inter-sensor relationships. Therefore, it generates a single, sparse, and interpretable graph per sample. Across multiple benchmark datasets (P12, P19, MIMIC-III, and PAM), WaveGNN delivers consistently strong performance, whereas other state-of-the-art baselines tend to perform well on some datasets or tasks but poorly on others. While WaveGNN does not necessarily surpass every method in every case, its consistency and robustness across diverse settings set it apart. Moreover, the learned graphs align well with known physiological structures, enhancing interpretability and supporting clinical decision-making.
\end{abstract}


\section{Introduction}
Continuous monitoring devices, electronic health‐record (EHR) systems, and wearable sensors now produce vast clinical time-series. Effectively modeling these datasets is crucial for downstream applications such as precision medicine, monitoring patient health, optimizing treatments, and managing disease outbreaks. Unlike the ideal, uniformly sampled sequences assumed in classical time-series analysis, clinical measurements often exhibit irregularities, caused by sensor malfunctions, different sampling rates across sensors, and cost-saving measures~\cite{zhang2021graph}. This variability in data collection renders traditional machine learning approaches, which rely on a fixed number of regularly sampled observations across all sensors, ineffective for accurate predictions~\cite{horn2020set, zhang2021graph}.

Understanding clinical time series requires capturing both the individual progression of each physiological signal and how different signals interact over time.  For example, tracking heart rate in isolation provides only partial information, whereas modeling its temporal relationship with blood pressure can reveal important clinical patterns. Incorporating such inter-series dependencies is essential for improving diagnostic accuracy.  Consequently, modelling these data poses two coupled challenges: (1) Intra-series irregularity, where each sensor records observations at uneven intervals and frequently experiences missing measurements. (2) Inter-series discrepancy, where different sensors operate at varying frequencies, resulting in misaligned, unequal-length sequences. These complexities significantly hinder traditional modeling methods, emphasizing the need for novel approaches capable of effectively addressing both types of irregularities.

A large body of prior work addresses intra-series irregularity. One line of methods applies interpolation or imputation to fill missing values, converting irregular series into regularly sampled sequences for downstream models~\cite{lipton2016directly,shukla2018interpolation,shukla2020multi}. While effective in some cases, these methods risk distorting temporal patterns and removing information carried by the structure of missingness itself. Alternative approaches such as GRU-D~\cite{che2018recurrent} and Neural ODEs~\cite{rubanova2019latent} avoid imputation by directly modeling irregular time intervals, but they typically operate on each sensor independently and do not account for inter-series interactions. This independence overlooks the fact that measurements from related sensors often exhibit strong dependencies (e.g., correlations between heart rate and respiration), so leveraging inter-series dynamics can help impute missing values more accurately and preserve meaningful temporal patterns. Without modeling such inter-series relationships, intra-series imputation risks being less reliable, particularly when data sparsity is severe.

Recent work has attempted to model inter-variable dependencies in irregular time series, but these approaches introduce key limitations. Many methods, such as UTDE~\cite{mcdermott2021comprehensive}, T-PatchGNN~\cite{zhang2024irregular}, and Warpformer~\cite{zhang2023warpformer}, bypass irregularity by first converting the entire irregular MTS into a regularly spaced representation. UTDE applies learned interpolation to fill in missing values, T-PatchGNN divides each signal into fixed-length patches, and Warpformer transforms signals onto a shared grid by downsampling high-frequency sensors and upsampling sparse ones. These transformations can distort temporal patterns and remove information carried by missingness. They also introduce additional hyperparameters such as patch size, stride, and grid resolution, which require tuning for each dataset. Moreover, these models do not produce a consistent or interpretable structure across variables, making it difficult to understand or trust the relationships they learn in critical domains like healthcare.

On the other hand, graphs have long been used to model structured and interpretable relationships. Recently, graph-based architectures have been adopted to explicitly model inter-variable dependencies in irregular time-series, aiming to address inter-series discrepancy. This is particularly desirable in healthcare, where understanding interactions between physiological signals is critical for decision-making~\cite{zhang2021graph}.
Recent graph-based techniques for modeling clinical time series, T-PatchGNN~\cite{zhang2024irregular} and GraFITi~\cite{yalavarthi2024grafiti}, often face critical limitations: they either construct multiple fragmented graphs that are difficult to aggregate into a coherent representation, or they restrict interactions between variables only through simultaneous observations, failing to model relationships effectively when sensor data are not aligned. Additionally, many of these methods do not produce an interpretable, unified graph structure, undermining transparency and limiting their clinical applicability. We compare with both models in our experiments and analyze their limitations in Section~\ref{sec:experiments}, alongside a detailed discussion in Section~\ref{sec:related-work}.

To address the limitations of existing approaches, we propose WaveGNN, a model that jointly captures intra-series irregularity and inter-series dependencies without converting the data to a regular format or introducing task-specific architectural hyperparameters\footnote{Code and data pointers are available at \url{https://github.com/USC-InfoLab/WaveGNN/}.}. WaveGNN combines a Transformer-based temporal encoder with a single dynamic graph neural network that operates over sensors, enabling direct modeling of both intra-series temporal patterns and inter-sensor relationships.

The Transformer-based encoder models each sensor independently. It uses masked attention to ignore missing values during computation and applies relative time encodings to capture the time gaps between observations. To account for the fact that recent measurements are often more relevant than older ones in clinical settings, the encoder includes a learnable decay mechanism that assigns lower weights to older inputs. This design allows the model to process irregular input sequences directly, without requiring imputation, interpolation, or resampling.

On top of the per-sensor representations, the dynamic graph module constructs a sparse, interpretable graph for each sample. Each sensor is treated as a node, and edges are formed based on two signals: (1) local similarity derived from Transformer representations within a time window and (2) global relationships encoded via learnable embeddings shared across all samples. A gating mechanism balances these short-term and long-term signals. The resulting graph reflects both sample-specific and population-level dependencies, and its structure is adapted dynamically for each sample. GNN layers propagate information across this graph to form a compact representation used for downstream classifications. Unlike prior models, WaveGNN produces a single, stable interaction graph per sample that is both clinically interpretable and robust to irregular sampling.

We evaluate WaveGNN on four real-world healthcare datasets covering binary, multi-class, 
and multi-label classification tasks. Across all benchmarks, WaveGNN demonstrates 
\emph{consistent performance}, achieving the lowest average rank (1.67) across 12 evaluations. The model is also highly 
\emph{robust} to missing data: across experiments simulating different levels of missingness, WaveGNN outperforms all baselines 
in 20 out of 24 experiments and ranks second in the remaining four, with relative F1 
improvements of up to 12.3\% on PAM. Ablation studies further show the necessity of 
jointly modeling inter- and intra-series dependencies, with F1 drops of 35.3\% and 15.1\% 
when these components are removed. Finally, WaveGNN produces interpretable graphs whose 
learned dependencies align with known physiological relationships 
(Section~\ref{sec:interpretability}), highlighting its utility for clinical decisions.

The remainder of the paper is organized as follows. Section~\ref{sec:related-work} reviews related work. Section~\ref{sec:prem} covers preliminaries and problem definition. Section~\ref{sec:meth} describes the WaveGNN architecture. Section~\ref{sec:experiments} presents experiments on four clinical datasets and analyzes model interpretability. Section~\ref{sec:conclusion} concludes the paper.


\section{Related Work}
\label{sec:related-work}
{\bf Modeling Intra-Series Irregularity}
Intra-series irregularity arises from variable sampling rates, missing values, and uneven observation intervals within each sensor. One line of work addresses this by converting irregular sequences into regularly spaced time series via interpolation. For instance, IP-Net~\cite{shukla2018interpolation} and DGM$^2$-O~\cite{wu2021dynamic} apply kernel interpolation over fixed reference points, while mTAND~\cite{shukla2020multi} combines multi-head interpolation with time-aware attention. These methods support standard architectures but risk introducing bias and losing information inherent in the pattern of missingness~\cite{mcdermott2021comprehensive,zhang2023improving}. An alternative direction aims to model irregular sequences directly. GRU-D~\cite{che2018recurrent} incorporates exponential decay terms into both inputs and hidden states to reflect temporal recency. SeFT~\cite{horn2020set} uses set functions to generate permutation-invariant embeddings from unordered observations. Neural ODEs~\cite{rubanova2019latent} model continuous-time dynamics to naturally handle irregular sampling. While these approaches preserve temporal structure, they typically process each sensor independently and do not model cross-variable relationships.

{\bf Modeling Inter-Series Dependencies}
Capturing dependencies across variables is critical in clinical time series, where interactions, such as between heart rate and blood pressure, can indicate physiological states. Several recent methods address this by building inter-variable representations on top of regularized input representation. UTDE~\cite{mcdermott2021comprehensive} jointly learns interpolation functions and applies cross-variable attention within a Transformer. Warpformer~\cite{zhang2023warpformer} resamples signals onto a shared temporal grid via learnable downsampling and upsampling. T-PatchGNN~\cite{zhang2024irregular} segments each signal into fixed-length regularly-spaced patches and constructs local graphs to capture short-term inter-series dependencies. However, these models rely on regularization that may distort temporal cues and introduce tuning overhead via additional hyperparameters (e.g., patch sizes, strides, or grid resolutions). Moreover, they do not yield consistent or interpretable inter-variable relationship structures.

{\bf Graph-Based Approaches}
Graph-based models offer a natural way to encode structured and interpretable dependencies in multivariate time series. Prior works that provide interpretable relationship structures include BysGNN~\cite{hajisafi2023learning}, which constructs context-aware dynamic graphs for spatiotemporal forecasting, and NeuroGNN~\cite{hajisafi2024dynamic}, which captures multi-channel EEG interactions for seizure detection. However, these models are designed for regularly sampled inputs.

To extend graph modeling to irregular settings, Raindrop~\cite{zhang2021graph} introduces event-based message passing over variable nodes. However, it lacks explicit intra-series modeling, treats all observations equally, and does not encode long-term dependencies between variables. T-PatchGNN~\cite{zhang2024irregular} builds patch-specific graphs, but these vary across time and do not produce a coherent interaction graph per instance. GraFITi~\cite{yalavarthi2024grafiti} constructs a bipartite graph over time and variable nodes, but restricts interactions to co-observed timestamps, limiting its effectiveness under heavy misalignment.

In contrast, WaveGNN integrates a decay-aware Transformer for intra-series encoding with a dynamic, per-sample graph over variables that captures both short-term temporal similarity and long-range structural dependencies. This yields a consistent, sparse, and interpretable interaction graph for each instance, without relying on interpolation, handcrafted temporal alignment, or task-specific hyperparameters. Evaluation is presented in Section~\ref{sec:experiments}.

\section{Preliminaries}
\label{sec:prem}






\begin{definition}[Time Series Observation]
A time series observation, \( s^{t}_{i, v} \), is a single measurement recorded by sensor \( v \) at timestamp \( t \) for sample \( i \). 
Sequential observations over time form a time series \( s_{i,v} \) for that sample and sensor.
\end{definition}

\begin{definition}[Irregular Multivariate Time Series Dataset]
Let \( D = \{(S_i, t_i, p_i, y_i)\}_{i=1}^{N} \) be a dataset where each \( S_i \) is an irregularly sampled multivariate time series for the \( i \)-th sample, \( t_i = \{ t_{i,v} \}_{v=1}^{n} \) are the corresponding timestamps, \( y_i \in \{1, \ldots, C\} \) is the associated categorical outcome label, with \( C \) representing the total number of classes, and \( p_i \) is an optional set of static features (e.g., patient demographics). Each \( S_i \) consists of multiple time series, one for each sensor \( v \). Specifically, \( S_i = \{ s_{i, v} \}_{v=1}^{n} \), where \( s_{i, v} \) is the time series of observations recorded by sensor \( v \) for sample \( i \). The timestamps in \( t_{i,v} \) are irregularly spaced, meaning the intervals between consecutive timestamps can vary.
\end{definition}


{\bf Problem Definition (Irregular Multivariate Time Series Classification Task)}
Given a dataset \( D = \{(S_i, t_i, p_i, y_i)\}_{i=1}^{N} \) of irregularly sampled multivariate time series, the goal is to learn a function \( f: S_i \rightarrow g_i \) that maps the sample \( s_i \), to a fixed-length representation \( g_i \). This representation can be utilized to predict a label \( \hat{y_i} \in \{1, \ldots, C\} \) relevant to the downstream task.




\begin{figure}[t]
\centering
    \begin{adjustbox}{width=0.5\textwidth, center}
        \includegraphics[]{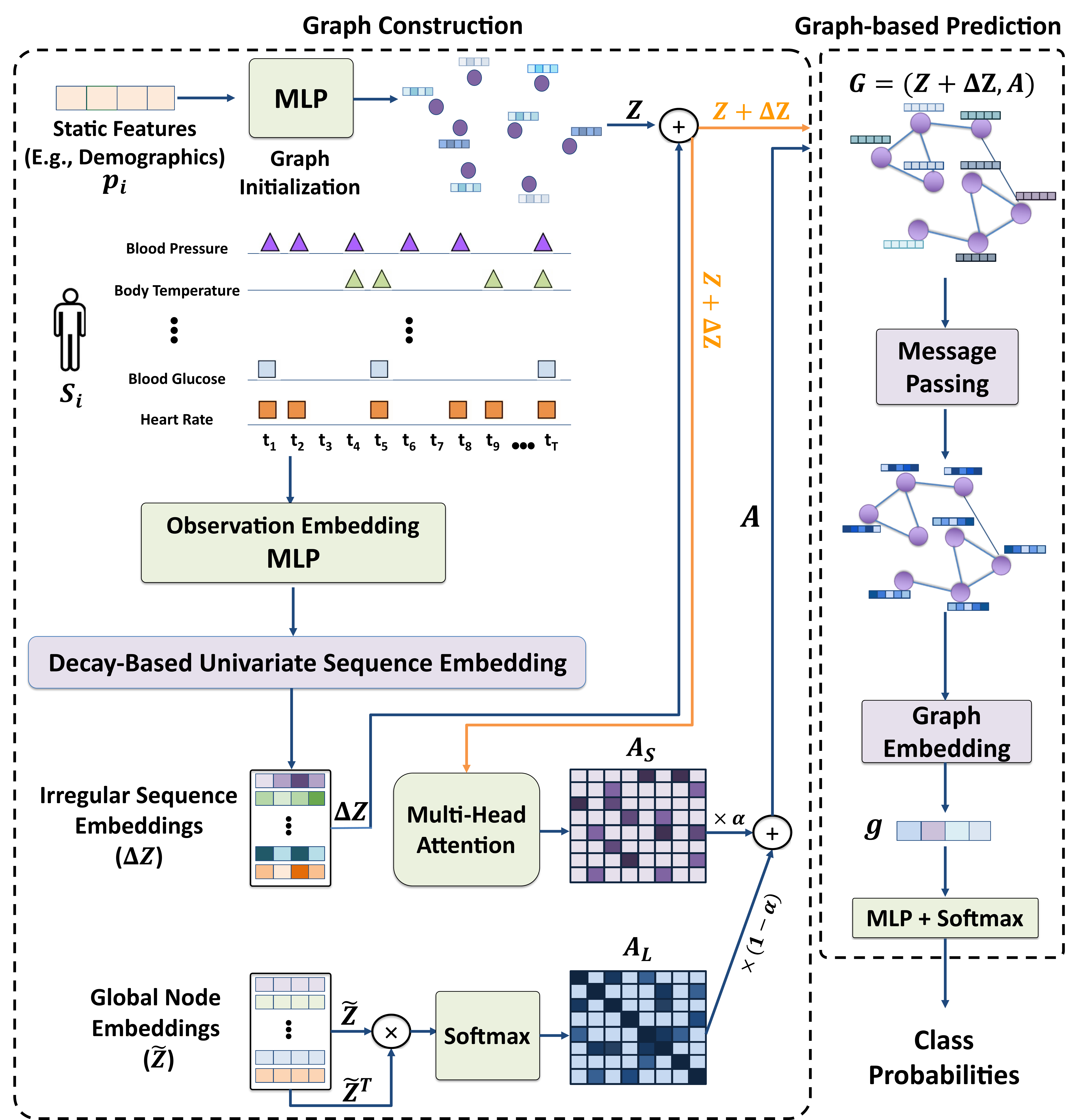}
    \end{adjustbox}
\caption{Overall WaveGNN Framework}
\label{fig:wavegnn-overall}
\end{figure}

\section{Methodology}\label{sec:meth}


Figure~\ref{fig:wavegnn-overall} presents the end-to-end pipeline of WaveGNN. In the first step, WaveGNN allocates nodes for each sensor, initializing their states using static features from the dataset, like patient demographics, if available, or randomly if not. Next, each sensor’s irregular observation sequence is embedded using a customized Transformer encoder. This encoder employs masked attention, modified temporal encoding, and a decay mechanism to handle irregularities and capture correlations within each sensor's sequence. The node states are updated based on these sequence embeddings to represent the latent state of each sensor after the observed window. In the third step, WaveGNN establishes edges that represent the correlations across sensors to effectively utilize inter-series correlations. This allows WaveGNN to compensate for missing observations during the input window by leveraging information from other related sensors to each sensor. These edges are formed based on the updated node states, capturing short-term relationships during the observed window, and on learned global node embeddings, capturing long-term relationships derived from the entire training dataset. 
Next, WaveGNN performs message passing, updating node states to incorporate information from inferred relationships. This allows each sensor to benefit from other sensors' data, reducing the impact of irregularity. Finally, the updated node states are aggregated into a graph-level embedding vector, which is fed into the prediction head. The detailed steps are presented below.

\subsection{Sensor Graph Construction}

\textbf{Initialization.} In WaveGNN, the graph initialization step establishes the preliminary graph structure without connecting nodes. These will later be updated to reflect the relationships and state of the given sample during the observed time window. The graph comprises nodes, each representing a different sensor measuring a time-dependent feature.

For each sample $i$, each node is initialized with a state vector derived from static features $p_i$ corresponding to the sample (e.g., demographics such as age and gender, and clinical contexts such as ICU types in medical downstream tasks). If unavailable, the vectors are initialized randomly. These features are transformed using a two-layer MLP with ReLU activations to create high-dimensional representations that encapsulate the sample's baseline characteristics and provide a sample-specific initial context for the task:
\begin{equation}
    Z_i = \text{MLP}(p_i)
\end{equation}
Here, $p_i$ represents the sample $i$'s static features and
$Z_i = \{z_{i,1}, \cdots z_{i,n}\}\in{\mathbb{R}}^{n\times M}$ represents the matrix of initial state vectors for all nodes where $M$ is the embedding dimension.

\noindent \textbf{Observation Embedding.} For sample $i$, for each sensor $v$, let the observation at time $t$ be denoted as $s_{i,v}^t$. To enhance the expressive power of our model~\cite{velivckovic2018graph}, each observation is mapped to a higher-dimensional space using an MLP with $\tanh$ activations:
\begin{equation}
    h_{i,v}^t = \text{MLP}(s_{i,v}^t)
\end{equation}
We use $\tanh$ activation to ensure that embeddings are centered around zero, which helps in capturing the inconsistencies in the observations. For instance, two observations that are inconsistent with each other can result in embeddings that are oppositely signed.

\noindent \textbf{Temporal Encoding.} To effectively capture the temporal dynamics and patterns within each sensor's time series, we augment the observation embeddings with time information. Inspired by GRU-D~\cite{che2018recurrent}, we similarly embed the relative timestamps that reflect the time difference between consecutive observations for each sensor.

To represent these time differences, we apply Time2Vec~\cite{kazemi2019time2vec}, which encodes the time information into a vector. Time2Vec includes two components: a sinusoidal component that captures periodic patterns, and a non-periodic transformation component that captures non-periodic patterns and time gaps. This dual encoding enables the model to incorporate both regular temporal cycles and irregular time gaps between observations. This turns the time intervals themselves into a valuable source of information rather than merely considering the relative position of observations, as done in the original Transformer~\cite{vaswani2017attention}.

For each observation at time $t_{v,j}$, where $v$ indicates the sensor and $j$ the index of the observation, we compute the time delta $\delta t_{v,j} = t_{v,j} - t_{v,j-1}$ as the difference between the current observation time and the previous observation time for that sensor. The time encoding vector $e_{\delta t_{v,j}}$ is then computed as:
\begin{equation}
e_{\delta t_{v,j}} = \text{Time2Vec}(\delta t_{v,j})
\end{equation}
Augmenting observation embeddings with $e_{\delta t_{v,j}}$ equips the model to better recognize and utilize the underlying temporal structure in the data.

\noindent \textbf{Decay-Based Univariate Sequence Embedding.}
In order to capture the intra-series correlations, we embed each irregular sequence of observations from each sensor separately. Given the sequence of $T$ observation embeddings $(h_{i,v}^{t_1}, \cdots, h_{i,v}^{t_T})\in \mathbb{R}^{T\times M}$ for sample $i$ and sensor $v$, we add the temporal encoding information to the observation embeddings. For simplicity, we omit the sample index $i$ in the following equations:
\begin{equation}
    h_v^{t_j} \leftarrow h_v^{t_j} + e_{\delta t_{v,j}} \quad \forall j \in \{1, 2, \dots, T\}
\end{equation}
Next, we pass the time-augmented sequence $H_v=(h_v^{t_1},\cdots ,h_v^{t_T})\in \mathbb{R}^{T\times M}$ through a Transformer encoder, using the mask vector $M_{v} = (m_{v, 1}, \cdots, m_{v,T}) \in \mathbb{R}^{T}$ as a padding mask. Here, each $m_{v,j}$ indicates whether there was an observation at time $t_j$ for sensor $v$ (1 if present, 0 if missing). If the observation is missing, the corresponding $h_v^{t_j}$ is a zero-filled place-holder vector. This way, the Transformer’s attention mechanism focuses only on the actual observations, effectively ignoring the missing ones while embedding the real observations:
\begin{equation}
    Z_v = \text{TransformerEncoder}(H_v, M_v)
\end{equation}

The output sequence $Z_v = (z_v^{t_1}, \cdots, z_v^{t_T})$ captures the intra-series correlations, with the attention mechanism ensuring that the presence of missing observations does not negatively affect the representation of the observed data.

Following this, WaveGNN utilizes a decay-based weighting scheme to aggregate the new sequence of embeddings into a single vector. This ensures that more recent observations have a greater influence on the final representation, as they provide the most up-to-date information about the sample. Moreover, WaveGNN's weighting accounts for the irregular time intervals between observations, considering that observations with shorter intervals between them should have similar impacts on the final embedding, whereas those with larger time gaps should be weighted differently.

To this end, WaveGNN first calculates the time differences of each observation from the most recent observation $t_{v,T}$:
\begin{equation}
    \delta t'_{v,j} = t_{v,T} - t_{v,j}
\end{equation}


Next, we compute an observation weight at $t_{v,j}$ as follows:
\begin{equation}
    w_{v,j} = m_{v,j}\cdot e^{-\eta \times \delta t'_{v,j}}
\end{equation}
where $\eta$ is a learned parameter that controls the decay rate. Mask value $m_{v,j}$ is multiplied so the missing observations are ignored. The exponential function is used because it provides a smooth decay of weights controlled by $\eta$ as the time differences increase, ensuring that observations further away from the most recent time have smaller weights.

Finally, we normalize the weights using a softmax function so that they sum up to 1:
\begin{equation}
    \tilde{w}_{v,j} = \frac{m_{v,j} \cdot e^{-\eta \times \delta t'_{v,j}} }{\sum_{k=1}^{T} m_{v,k} \cdot e^{-\eta \times \delta t'_{v,k}}}
\end{equation}

We then aggregate the embeddings to obtain a single vector representation for the sequence using the calculated weights:
\begin{equation}
    \Delta z_v = \sum_{j=1}^{T} \tilde{w}_{v,j} \cdot z_v^{t_j}
\end{equation}
The final embedding vector $\Delta z_v$ effectively captures the intra-series dependencies and the importance of each observation in sensor $v$. This representation is then used to update the node state vector for the sensor:
\begin{equation}
z_{i,v} \leftarrow z_{i,v} + \Delta z_v
\end{equation}
This update is applied in a residual manner to enhance the gradient flow during training and allow the model to refine the initial context derived from static features with information from the actual observations.

\noindent \textbf{Inferring Inter-series Relationships.}
Now that node states are updated, WaveGNN infers the inter-node relationships by combining two distinct types of adjacency matrices, each representing the relationships from a different perspective:

\hspace*{5pt} \textbf{$\bullet$ Dynamic Short-Term Dependencies ($A_S$)}: This matrix is derived from the multi-head attention mechanism~\cite{vaswani2017attention} applied to the updated node states. It captures the temporal similarity across node states of different sensors within the observed window, reflecting immediate, short-term dependencies among the nodes. This approach has been shown to be effective in capturing dynamic node relationships~\cite{velivckovic2017graph,hajisafi2023learning,hajisafi2024dynamic}.
    \begin{equation}
    A_S = \text{Multihead\_Attention}(\text{keys}=Z, \text{queries}=Z)
    \end{equation}

\hspace*{5pt} \textbf{$\bullet$ Static Long-Term Similarities ($A_L$)}: WaveGNN learns global node embeddings $\tilde{Z} = \{\tilde{z}_1, \dots, \tilde{z}_n\} \in \mathbb{R}^{n \times M}$, where $M$ is the embedding dimension. These embeddings reflect long-term similarities between sensors, learned from the entire dataset for a specific downstream task. $A_L$ matrix is then computed using the dot product between these global node embeddings, followed by a softmax normalization:
    \begin{equation}
    A_L = \text{softmax}(\tilde{Z} \times \tilde{Z}^T)
    \end{equation}

The final adjacency matrix $A$ is formed by combining the two similarity matrices using a learnable parameter $\alpha$:
\begin{equation}\label{eq:adj}
    A = \alpha A_S + (1-\alpha) A_L
\end{equation}

This allows the model to learn which of these similarities are more important in the downstream task. Finally, we zero out the smallest $K\%$ of the elements in $A$ to make the resulting adjacency matrix sparser.

\subsection{Graph-based Prediction}
With the updated node states $Z$ and adjacency matrix $A$, the graph $G=(Z,A)$ for the given window of observations is constructed, encapsulating the latest node states and their inter-relationships. We perform GNN-based message passing on this graph to capture the node relationships through the node embeddings. Afterward, we apply pooling and transformations to obtain a single embedding vector representing all important node embeddings. The final embedding is used to make the prediction.

\noindent \textbf{Message Passing.} To perform message passing, the graph is passed through a GNN block that utilizes a modified Graph Convolutional Networks (GCN) variant~\cite{kipf2017semisupervised}. In this variant, we remove the normalization term and add residual connections between the message-passing layers to preserve directed relationships in the graph and mitigate oversmoothing~\cite{chen2020measuring}, respectively. This yields the node embeddings matrix $V \in \mathbb{R}^{n \times M'}$, where $M'$ represents the embedding dimension.

\noindent \textbf{Graph Embedding.} To derive a graph-level representation from the node embeddings, two operations are performed: 

    \hspace*{5pt} \textbf{$\bullet$ Max Pooling}: Applied to the node embeddings to extract the most significant features, helping to capture the most expressive node attributes relevant to the prediction task.
    
    \hspace*{5pt} \textbf{$\bullet$ Transformation}: All node embeddings are concatenated into a single vector, which is then transformed via an MLP to integrate information across all nodes.

\noindent \textbf{Final Prediction.} The final $l$-dimensional graph-level representation $g\in \mathbb{R}^l$ is obtained by first concatenating the outputs of the max pooling layer and the MLP-transformed concatenated embeddings, followed by an MLP layer for dimensionality reduction.

Finally, the graph-level embedding $g$ is utilized for the downstream task of interest. In our approach, we pass this through another MLP to predict the label of the samples:

\begin{equation}
\hat{y} = MLP(g)
\end{equation}

Where $\hat{y}$ represents the predicted class probabilities.

\section{Experiments}
\label{sec:experiments}
\subsection{Experimental Setup}
\textbf{Datasets and Evaluation Setup.}
We evaluate WaveGNN on four real-world healthcare datasets. \textbf{P12}~\cite{goldberger2000physiobank} and \textbf{MIMIC III}~\cite{johnson2016mimic} contain ICU patient data, with labels indicating survival during hospitalization (48-IHM). {MIMIC III} also includes labels for multilabel phenotype classification (24-PHE). \textbf{P19}~\cite{reyna2020early} focuses on sepsis detection, while \textbf{PAM}~\cite{reiss2012introducing} is an activity monitoring dataset with labels for different activities. P12, P19, and MIMIC III are naturally irregular datasets, while irregularity in PAM is introduced artificially. The dataset statistics' are presented in Table~\ref{tab:dataset_stats}. For a fair comparison, we followed the same preprocessing steps with previous studies~\cite{zhang2021graph, harutyunyan2019multitask}, and adopted the same dataset splits. To that extent, the training, validation, and testing ratios for P12, P19, and PAM is 80:10:10, and 85:7.5:7.5 for MIMIC III. The indices of these splits are fixed across all methods and experiments. Evaluation metrics also match those used in prior studies to maintain consistency, and all hyperparameters were kept the same across all datasets and experiments. 


\begin{table}[h]
    \centering
    \caption{Statistics of the datasets used in experimental study.}
    \label{tab:dataset_stats}
    \scalebox{0.85}{
    \begin{tabular}{ccccc}
        \toprule
        \textbf{Statistics} & {MIMIC III} & {P12} & {P19} & {PAM} \\
        \midrule
        \# of samples & 16,143 & 11,988 & 38,803 & 5,333\\
        \# of timestamps & { $\leq$} 700 & 215 & 60 & 600\\
        \# of sensors & 17 & 36& 34& 17 \\
        static info & False & True & True & False\\
        missing ratio (\%) & 72.07 & 88.40 & 94.90 & 60.00 \\
        imbalanced & True & True & True & False \\
        \# of classes & 2 (IHM) / 25 (PHE) & 2 & 2 & 8 \\
        \bottomrule
    \end{tabular}
    }
\end{table}

\noindent \textbf{Baselines.} We compare WaveGNN to 11 methods. \textit{IP-Net}~\cite{shukla2018interpolation}, \textit{DGM$^2$-O}~\cite{wu2021dynamic}, \textit{mTAND}~\cite{shukla2020multi}, and \textit{UTDE}~\cite{zhang2023improving} are interpolation-based models. \textit{GRU-D}~\cite{che2018recurrent}, \textit{SeFT}~\cite{horn2020set}, \textit{MTGNN}~\cite{wu2020connecting}, and \textit{WarpFormer}~\cite{zhang2023warpformer} are specialized architectures for modeling irregular time series. \textit{Raindrop}~\cite{zhang2021graph}, \textit{T-PatchGNN}~\cite{zhang2024irregular}, and \textit{GraFITi}~\cite{yalavarthi2024grafiti} are graph-based methods that model inter-variable relationships under irregular sampling. Among these, GraFITi and WarpFormer represent recent state-of-the-art models.

\noindent \textbf{Dataset Preprocessing.}\label{app:data-preproc}
To ensure compatibility with previous approaches, we followed the preprocessing steps outlined in~\cite{harutyunyan2019multitask}~\footnote{\tiny{\url{https://github.com/YerevaNN/mimic3-benchmarks/tree/master/mimic3benchmark}}} for the MIMIC III dataset. For the P12, P19, and PAM datasets, we used the preprocessed data provided by~\cite{zhang2021graph}~\footnote{\tiny{\url{https://figshare.com/articles/dataset/P19_dataset_for_Raindrop/19514338/1?file=34683070}}}~\footnote{\tiny{\url{https://figshare.com/articles/dataset/P12_dataset_for_Raindrop/19514341/1?file=34683085}}}~\footnote{\tiny{\url{https://figshare.com/articles/dataset/PAM_dataset_for_Raindrop/19514347/1?file=34683103}}}. To compare our approach with baselines, we used the same data split indexes as provided by \cite{zhang2021graph} for P12, P19, and PAM, and by~\cite{harutyunyan2019multitask} for MIMIC.

\noindent \textbf{Computational Resources}\label{app:hardware}
Our experiments were performed on a cluster node equipped with a 24 GB NVIDIA GeForce RTX 3090 GPU with CUDA version 11.8 and a Ryzen 9 5900 12-core processor, running on Ubuntu 20.04. Furthermore, all neural network models are implemented based on PyTorch version 2.2.2 and Pytorch Geometric version 2.5.3 on Python version 3.9.19.

\noindent \textbf{Hyperparameter Configuration}\label{app:hyperparam}
\textbf{Training Setup.} All experiments were run for 20 epochs using AdamW~\cite{loshchilov2017decoupled} with a learning rate of 1e-4, dropout rate of 0.1, and weight decay of 1e-5. A batch size of 4 was used with gradient accumulation over 32 steps to achieve an effective batch size of 128. We employed a ReduceLROnPlateau scheduler (factor 0.1, patience 1, min LR 1e-8). Model selection was based on AUROC (binary tasks), weighted F1 (PAM), and AUPRC (24-PHE), with early stopping (patience 10). Each experiment was repeated 3 times, and mean and standard deviation were reported.

\textbf{Input Window.} Full observation windows were used for P12 (max 215), P19 (60), and PAM (600). For MIMIC, we used the last 700 timestamps to ensure memory efficiency.

\textbf{WaveGNN Configuration.} Static features (when available) were embedded via a 2-layer MLP with ReLU activations to $Z_i \in \mathbb{R}^{128}$. Observation embeddings $h_{i,v}^t \in \mathbb{R}^{64}$ were generated via a 3-layer MLP with $\tanh$ activations. Sensor sequences were encoded using a 2-layer Transformer with 64-dim outputs, using 64-dim relative time encodings and an initial decay parameter $\eta=0.1$. A linear projection then mapped outputs to 128-dim.

Short-term sensor similarity ($A_S$) was computed using 8-head self-attention. Long-term context came from a learnable global embedding matrix $\tilde{Z} \in \mathbb{R}^{\#sensors \times 128}$; $\alpha$ was initialized to 0.5 to balance $A_S$ and $A_L$. To maintain sparsity, we pruned the lowest 30\% of edges.

The GNN module consisted of 3 GCN layers with residual connections, layer norm, and output dimension 128. The graph embedding module concatenated (1) a 3-layer MLP-transformed global node embedding and (2) a feature-wise max pooling over all nodes, yielding a 256-dim vector. This was passed through a final 2-layer MLP with ReLU to produce the task-specific output.

Additional hyperparameter details are provided in the code readme.

\begin{table*}[h]
    \centering
    \caption{Performance comparison with the state-of-the-art on P12, P19, and PAM datasets. P12 and P19 are evaluated on mortality and sepsis prediction, respectively, based on the AUROC and AUPRC metrics. The PAM dataset is evaluated on activity classification in terms of (weighted) Accuracy, Precision, Recall, and F1 Score. The reported values are averaged across 3 independent runs. \textbf{Best} and \underline{second best} values are highlighted.}
    \label{tab:overall_results_1}
    \begin{tabular}{lcccccccccc}
        \toprule 
        \textbf{Methods} & \multicolumn{2}{c}{\textbf{P12}} & \multicolumn{2}{c}{\textbf{P19}} & \multicolumn{4}{c}{\textbf{PAM}} \\
        \cmidrule(lr){2-3} \cmidrule(lr){4-5} \cmidrule(lr){6-9}
         & {AUROC} & {AUPRC} & {AUROC} & {AUPRC} & {Accuracy} & {Precision} & {Recall} & {F1 score} \\
        \midrule
        GRU-D & 81.9 {$\pm$ 3.6} & 46.1 {$\pm$ 4.7} & 83.9 {$\pm$ 1.7} & 46.9 {$\pm$ 2.1} & 83.3 {$\pm$ 1.6} & 84.6 {$\pm$ 1.2} & 85.2 {$\pm$ 1.6} & 84.8 {$\pm$ 1.2} \\
        SeFT & 73.9 {$\pm$ 2.5} & 31.1 {$\pm$ 4.1} & 78.7 {$\pm$ 2.4} & 31.1 {$\pm$ 2.8} & 67.1 {$\pm$ 2.2} & 70.0 {$\pm$ 2.4} & 68.2 {$\pm$ 1.5} & 68.5 {$\pm$ 1.8} \\
        mTAND & \underline{84.2 {$\pm$ 0.8}} & \underline{48.2 {$\pm$ 3.4}} & 80.4 {$\pm$ 1.3} & 32.4 {$\pm$ 1.8} & 74.6 {$\pm$ 4.3} & 74.3 {$\pm$ 4.0} & 79.5 {$\pm$ 2.8} & 76.8 {$\pm$ 3.4} \\
        IP-Net & 82.6 {$\pm$ 1.4} & 47.6 {$\pm$ 3.1} & 84.6 {$\pm$ 1.3} & 38.1 {$\pm$ 3.7} & 74.3 {$\pm$ 3.8} & 75.6 {$\pm$ 2.1} & 77.9 {$\pm$ 2.2} & {76.6 {$\pm$ 2.8}} \\
        DGM$^2$-O & \textbf{84.4 {$\pm$ 1.6}} & {47.3 {$\pm$ 3.1}} & 86.7 {$\pm$ 3.4} & 44.7 {$\pm$ 11.7} & 82.4 {$\pm$ 2.3} & 85.2 {$\pm$ 1.2} & 83.9 {$\pm$ 2.3} & 84.3 {$\pm$ 1.8} \\
        MTGNN & 74.4 {$\pm$ 6.7} & 35.5 {$\pm$ 6.0} & 81.9 {$\pm$ 6.2} & 39.9 {$\pm$ 8.9} & 83.4 {$\pm$ 1.9} & 85.2 {$\pm$ 1.7} & 86.1 {$\pm$ 1.9} & 85.9 {$\pm$ 2.4} \\
        {Raindrop} & {82.8 {$\pm$ 1.7}} & {44.0 {$\pm$ 3.0}} & {87.0 {$\pm$ 2.3}} & {51.8 {$\pm$ 5.5}} & {88.5 {$\pm$ 1.5}} & {89.9 {$\pm$ 1.5}} & {89.9 {$\pm$ 0.6}} & {89.8 {$\pm$ 1.0}} \\
        {UTDE} & {81.1 {$\pm$ 2.2}} & {45.9 {$\pm$ 2.1}} & {84.3 {$\pm$ 3.1}} & {41.1 {$\pm$ 6.0}} & {90.4 {$\pm$ 1.8}} & {90.7 {$\pm$ 1.8}} & {90.4 {$\pm$ 1.8}} & {90.4 {$\pm$ 1.8}} \\
        T-PatchGNN & 62.1 {$\pm$ 4.8} & 21.3 {$\pm$ 1.8} & 82.0 {$\pm$ 3.2} & 52.1 {$\pm$ 2.3} & 90.6 {$\pm$ 1.6} & 90.9 {$\pm$ 1.8} & 90.6 {$\pm$ 1.6} & 90.6 {$\pm$ 1.7} \\
        Warpformer & 81.9 {$\pm$ 2.8} & 39.6 {$\pm$ 1.6} & \underline{89.4 {$\pm$ 0.3}} & \underline{64.9 {$\pm$ 0.4}} & 91.4 {$\pm$ 1.4} & 91.5 {$\pm$ 1.3} & 91.4 {$\pm$ 1.4} & 91.4 {$\pm$ 1.4} \\
        GraFITi & 80.3 {$\pm$ 1.0} & 37.7 {$\pm$ 0.4} & \textbf{90.5 {$\pm$ 1.1}} & \textbf{66.0 {$\pm$ 1.7}} & \underline{94.6 {$\pm$ 0.5}} & \underline{94.7 {$\pm$ 0.5}} & \underline{94.6 {$\pm$ 0.5}} & \underline{94.6 {$\pm$ 0.5}} \\
        \textbf{WaveGNN} & {83.9 {$\pm$ 1.2}} & \textbf{49.4 {$\pm$ 1.5}} & {88.0 { $\pm$ 0.9}} & {57.1 { $\pm$ 4.7}} & \textbf{95.6 {$\pm$ 1.1}} & \textbf{95.7 {$\pm$ 1.1}} & \textbf{95.6 {$\pm$ 1.1}} & \textbf{95.6 {$\pm$ 1.1}} \\
        \bottomrule
    \end{tabular}%
\end{table*}

\begin{table}[h]
    \centering
    \caption{Performance comparison with the state-of-the-art on MIMIC III dataset, for 48-IHM and 24-PHE tasks. The performance of 48-IHM is measured on F1 and AUPRC, and 24-PHE on F1 (Macro) and AUROC. We report the average results across 3 independent runs, highlighting \textbf{best} and \underline{second best} values.}
    \label{tab:mimic_results}
    \resizebox{0.5\textwidth}{!}{%
        \begin{tabular}{lcccc}
            \toprule 
            \textbf{Methods} & \multicolumn{2}{c}{\textbf{48-IHM}} & \multicolumn{2}{c}{\textbf{24-PHE}} \\
            \cmidrule(lr){2-3} \cmidrule(lr){4-5} 
            & {F1} & {AUPRC} & {F1} & {AUROC} \\
            \midrule
            GRU-D & 42.8 {$\pm$ 0.6} & 45.9 {$\pm$ 0.4} & 19.0 {$\pm$ 1.0} & 73.3 {$\pm$ 0.1} \\
            SeFT & 16.5 {$\pm$ 8.6} & 23.9 {$\pm$ 0.5} & 6.1 {$\pm$ 0.2} & 65.7 {$\pm$ 0.1} \\
            mTAND & 43.9 {$\pm$ 0.5} & 47.5 {$\pm$ 1.3} & 19.9 {$\pm$ 0.4} & 73.5 {$\pm$ 0.1} \\
            IP-Net & 37.2 {$\pm$ 2.8} & 39.4 {$\pm$ 1.1} & 17.9 {$\pm$ 0.7} & 73.5 {$\pm$ 0.1} \\
            DGM$^2$-O & 39.1 {$\pm$ 1.5} & 37.8 {$\pm$ 1.5} & 18.4 {$\pm$ 0.2} & 71.7 {$\pm$ 0.2} \\
            MTGNN & 38.6 {$\pm$ 2.5} & 36.5 {$\pm$ 2.1} & 14.5 {$\pm$ 1.7} & 70.6 {$\pm$ 0.7} \\
            {Raindrop} & {39.5 {$\pm$ 3.7}} & {36.2 {$\pm$ 0.4}} & {21.8 {$\pm$ 1.7}} & {74.0 {$\pm$ 0.9}} \\
            {UTDE} & \underline{45.3 {$\pm$ 0.7}} & \textbf{49.6 {$\pm$ 1.0}} &  \underline{24.9 {$\pm$ 0.4}} & \textbf{75.6 {$\pm$ 0.2}} \\
            T-PatchGNN & 39.8 {$\pm$ 1.4} & 36.1 {$\pm$ 0.8} & 15.7 {$\pm$ 1.6} & 70.9 {$\pm$ 0.9} \\
            Warpformer & 21.8 {$\pm$ 6.3} & 32.6 {$\pm$ 1.5} & 19.3 {$\pm$ 0.7} & 74.1 {$\pm$ 0.2} \\
            GraFITi & 28.8 {$\pm$ 1.1} & 37.1 {$\pm$ 1.1} & 15.5 {$\pm$ 0.7} & 72.7 {$\pm$ 0.2} \\

            \textbf{WaveGNN} & \textbf{46.5 {$\pm$ 0.9}} & \underline{47.8 {$\pm$ 1.3}} &  \textbf{30.8 {$\pm$ 0.2}} & \underline{74.9 {$\pm$ 0.1}} \\
        \bottomrule
    \end{tabular}
    }
\end{table}

\begin{table}[t]
    \centering
    \caption{Cross-dataset/task consistency across all 12 dataset-metric evaluation pairs
    (P12, P19, PAM, and MIMIC). We report average rank (lower is better), 
    the number of times each method ranked first (Top-1), and the number 
    of times each method ranked among the top two (Top-2).}
    \label{tab:overall-res-summary}
    \begin{tabular}{lccc}
        \toprule
        Method & Avg Rank & Top-1 Count & Top-2 Count \\
        \midrule
        WaveGNN      & \textbf{1.67} & \textbf{7} & \textbf{9} \\
        UTDE         & 4.58 & 2 & 4 \\
        GraFITi      & 5.25 & 2 & 6 \\
        Warpformer   & 5.04 & 0 & 2 \\
        Raindrop     & 5.50 & 0 & 0 \\
        DGM2-O       & 6.71 & 1 & 1 \\
        GRU-D        & 6.62 & 0 & 0 \\
        mTAND        & 6.88 & 0 & 2 \\
        T-PatchGNN   & 7.25 & 0 & 0 \\
        IP-Net       & 7.88 & 0 & 0 \\
        MTGNN        & 8.79 & 0 & 0 \\
        SeFT         & 11.83 & 0 & 0 \\
        \bottomrule
    \end{tabular}
    \vspace{-12pt}
\end{table}



\begin{table*}[t]
    \centering
    \caption{Robustness experiments on the PAM dataset under different missing sensor ratios and different settings. We report the average results across 3 independent runs. \textbf{Best} and \underline{second best} values are highlighted.}
    \label{tab:pam_results_2_3}
    \resizebox{\textwidth}{!}{
    \begin{tabular}{c c|c c c c|c c c c}
        \toprule
        \multicolumn{2}{c|}{\textit{Experiment Setting}} & \multicolumn{4}{c|}{\textit{Leave-fixed-sensors-out}} & \multicolumn{4}{c}{\textit{Leave-random-sensors-out}} \\
        {\textbf{Missing sensor ratio}} & {\textbf{Methods}} & \textbf{Accuracy} & \textbf{Precision} & \textbf{Recall} & \textbf{F1 score} & \textbf{Accuracy} & \textbf{Precision} & \textbf{Recall} & \textbf{F1 score} \\
        \midrule
        {10\%} 
        & GRU-D & 65.4 $\pm$ { 1.7} & 72.6 $\pm$ { 2.6} & 64.3 $\pm$ { 5.3} & 63.6 $\pm$ { 0.4} & 68.4 $\pm$ { 3.7} & 74.2 $\pm$ { 3.0} & 70.8 $\pm$ { 4.2} & 72.0 $\pm$ { 3.7} \\
        & SeFT & 58.9 $\pm$ { 2.3} & 62.5 $\pm$ { 1.8} & 59.6 $\pm$ { 2.6} & 59.6 $\pm$ { 2.6} & 40.0 $\pm$ { 1.9} & 40.8 $\pm$ { 3.2} & 41.0 $\pm$ { 0.7} & 39.9 $\pm$ { 1.5} \\
        & mTAND & 58.8 $\pm$ { 2.7} & 59.5 $\pm$ { 5.3} & 64.4 $\pm$ { 2.9} & 61.8 $\pm$ { 4.1} & 53.4 $\pm$ { 2.0} & 54.8 $\pm$ { 2.7} & 57.0 $\pm$ { 1.9} & 55.9 $\pm$ { 2.2} \\
        & RAINDROP & {77.2 $\pm$ { 2.1}} & {82.3 $\pm$ { 1.1}} & {78.4 $\pm$ { 1.9}} & {75.2 $\pm$ { 3.1}} & {76.7 $\pm$ { 1.8}} & \underline{79.9 $\pm$ { 1.7}} & {77.9 $\pm$ { 2.3}} & \underline{78.6 $\pm$ { 1.8}} \\
        & UTDE & {71.8 $\pm$ { 1.7}} & {76.5 $\pm$ { 2.6}}  & {71.8 $\pm$ { 1.7}}  & {70.6 $\pm$ { 2.3}} & {74.5 $\pm$ { 1.9}} & {75.9 $\pm$ { 2.1}} & {74.5 $\pm$ { 1.9}} & {74.6 $\pm$ { 2.1}} \\
        & GraFITi & 73.3 {$\pm$ 3.2} & 75.0 {$\pm$ 2.0} & 73.3 {$\pm$ 3.2} & 72.6 {$\pm$ 2.8} & 75.3 {$\pm$ 1.5} & 75.7 {$\pm$ 2.1} & 75.3 {$\pm$ 1.5} & 75.2 {$\pm$ 1.8} \\
        & Warpformer & \underline{86.3 {$\pm$ 1.4}} & \underline{86.7 {$\pm$ 1.8}} & \underline{86.3 {$\pm$ 1.4}} & \underline{86.3 {$\pm$ 1.4}} & \underline{78.8 {$\pm$ 1.6}} & 78.4 {$\pm$ 2.1} & \underline{78.8 {$\pm$ 1.6}} & 78.4 {$\pm$ 1.8} \\
        & T-PatchGNN & 63.8 {$\pm$ 4.5} & 61.5 {$\pm$ 8.6} & 63.8 {$\pm$ 4.5} & 61.8 {$\pm$ 6.4} & 65.2 {$\pm$ 2.1} & 66.8 {$\pm$ 1.8} & 65.2 {$\pm$ 2.1} & 64.0 {$\pm$ 2.2} \\

        & WaveGNN & \textbf{86.9 $\pm$ { 2.2}} & \textbf{88.0 $\pm$ { 1.3}} & \textbf{86.9 $\pm$ { 2.2}} &  \textbf{86.9 $\pm$ { 3.3}} & \textbf{82.2 $\pm$ { 1.5}} & \textbf{83.9 $\pm$ { 0.8}} & \textbf{82.2 $\pm$ { 1.5}} & \textbf{82.5 $\pm$ { 1.3}} \\
        \cmidrule{2-10}
        & \textit{Improvement} &  +0.69\% &  +1.50\% & +0.70\% &  +0.70\% &  +4.31\% &  +5.00\% & +4.31\% & +4.96\% \\
        \midrule \midrule
        {30\%} 
        & GRU-D & 45.1 $\pm$ { 2.9} & 51.7 $\pm$ { 6.2} & 42.1 $\pm$ { 6.6} & 47.2 $\pm$ { 3.9} & 58.0 $\pm$ { 2.0} & 63.2 $\pm$ { 1.7} & 58.2 $\pm$ { 3.1} & 59.3 $\pm$ { 3.5} \\
        & SeFT & 32.7 $\pm$ { 2.3} & 27.9 $\pm$ { 2.4} & 34.5 $\pm$ { 3.0} & 28.0 $\pm$ { 1.4} & 31.7 $\pm$ { 1.5} & 31.0 $\pm$ { 2.7} & 32.0 $\pm$ { 1.2} & 28.0 $\pm$ { 1.6} \\
        & mTAND & 27.5 $\pm$ { 4.5} & 31.2 $\pm$ { 7.3} & 30.6 $\pm$ { 4.0} & 30.8 $\pm$ { 5.6} & 34.7 $\pm$ { 5.5} & 43.4 $\pm$ { 4.0} & 36.3 $\pm$ { 4.7} & 39.5 $\pm$ { 4.4} \\
        & RAINDROP & {52.4 $\pm$ { 2.8}} & {60.9 $\pm$ { 3.8}} & {51.3 $\pm$ { 1.6}} & {48.4 $\pm$ { 5.6}} & {60.3 $\pm$ { 3.3}} & \underline{68.1 $\pm$ { 4.6}} & {60.3 $\pm$ { 3.3}} & {60.3 $\pm$ { 3.3}} \\
        & UTDE & {41.2 $\pm$ { 2.9}} & {46.9 $\pm$ { 6.5}} & {41.2 $\pm$ { 2.9}} & {35.7 $\pm$ { 3.9}} & 57.9 {$\pm$ 2.5} & 65.0 {$\pm$ 3.4} & 57.9 {$\pm$ 2.5} & 58.4 {$\pm$ 2.8} \\
        & GraFITi & 59.7 {$\pm$ 4.3} & 64.2 {$\pm$ 5.6} & 59.7 {$\pm$ 4.3} & 58.9 {$\pm$ 5.2} & 63.5 {$\pm$ 5.9} & 66.2 {$\pm$ 5.2} & 63.5 {$\pm$ 5.9} & \underline{63.9 {$\pm$ 5.9}} \\
        & Warpformer & \underline{64.8 {$\pm$ 5.6}} & \underline{67.5 {$\pm$ 4.1}} & \underline{64.8 {$\pm$ 5.6}} & \underline{62.6 {$\pm$ 5.7}} & \underline{65.3 {$\pm$ 4.1}} & 64.5 {$\pm$ 4.1} & \underline{65.3 {$\pm$ 4.1}} & 63.6 {$\pm$ 5.0} \\
        & T-PatchGNN & 43.9 {$\pm$ 3.0} & 36.8 {$\pm$ 5.1} & 43.9 {$\pm$ 3.0} & 36.9 {$\pm$ 3.3} & 59.8 {$\pm$ 1.5} & 65.4 {$\pm$ 1.0} & 59.8 {$\pm$ 1.5} & 57.7 {$\pm$ 1.6} \\

        & WaveGNN & \textbf{71.2 $\pm$ { 2.8}} & \textbf{75.7 $\pm$ { 3.4}} & \textbf{71.2 $\pm$ { 2.8}} & \textbf{70.3 $\pm$ { 3.2}} & \textbf{68.6 $\pm$ { 1.6}} & \textbf{75.1 $\pm$ { 1.8}} & \textbf{68.6 $\pm$ { 1.6}} & \textbf{69.0 $\pm$ { 1.3}} \\
        \cmidrule{2-10}
        & \textit{Improvement} & +9.87\% & +12.14\% & +9.87\% & +12.30\% & +5.05\% & +10.27\% & +5.05\% & +7.98\% \\
        \midrule \midrule
        {50\%} 
        & GRU-D & 37.3 $\pm$ { 2.7} & 29.6 $\pm$ { 5.9} & 32.8 $\pm$ { 4.6} & 26.6 $\pm$ { 5.9} & {49.7 $\pm$ { 1.2}} & 52.4 $\pm$ { 0.3} & 42.5 $\pm$ { 1.7} & {47.5 $\pm$ { 1.2}} \\
        & SeFT & 24.7 $\pm$ { 1.1} & 15.9 $\pm$ { 2.7} & 25.3 $\pm$ { 2.6} & 18.2 $\pm$ { 2.4} & 26.4 $\pm$ { 1.4} & 23.0 $\pm$ { 2.9} & 27.5 $\pm$ { 0.4} & 23.5 $\pm$ { 1.8} \\
        & mTAND & 16.9 $\pm$ { 3.1} & 12.6 $\pm$ { 5.5} & 17.0 $\pm$ { 1.6} & 13.9 $\pm$ { 4.0} & 20.9 $\pm$ { 3.1} & 35.1 $\pm$ { 6.1} & 23.0 $\pm$ { 3.2} & 27.7 $\pm$ { 3.9} \\
        & RAINDROP & \textbf{{46.6 $\pm$ { 2.6}}} & \underline{{44.5 $\pm$ { 2.6}}} & {{42.4 $\pm$ { 3.9}}} & \underline{{38.0 $\pm$ { 4.0}}} & {47.2 $\pm$ { 4.4}} & \underline{59.4 $\pm$ { 3.9}} & {44.8 $\pm$ { 3.5}} & {47.6 $\pm$ { 5.2}} \\
        & UTDE & {30.6 $\pm$ { 2.4}} & {32.9 $\pm$ { 13.5}} & {30.6 $\pm$ { 2.4}} & {25.1 $\pm$ { 3.9}} & {43.5 $\pm$ { 1.7}} & {52.7 $\pm$ { 4.0}} & {43.5 $\pm$ { 1.7}} & {43.2 $\pm$ { 1.9}} \\
        & GraFITi & 22.9 {$\pm$ 1.2} & 10.6 {$\pm$ 0.9} & 22.9 {$\pm$ 1.2} & 12.4 {$\pm$ 1.9} & 47.9 {$\pm$ 3.4} & 53.4 {$\pm$ 4.4} & 47.9 {$\pm$ 3.4} & 48.4 {$\pm$ 4.5} \\
        & Warpformer & 43.6 {$\pm$ 3.2} & 35.3 {$\pm$ 3.4} & \textbf{43.6 {$\pm$ 3.2}} & 35.0 {$\pm$ 2.7} & \textbf{53.6 {$\pm$ 2.1}} & 54.0 {$\pm$ 2.4} & \textbf{53.6 {$\pm$ 2.1}} & \underline{50.7 {$\pm$ 3.4}} \\
        & T-PatchGNN & 31.0 {$\pm$ 1.7} & 15.6 {$\pm$ 6.0} & 31.0 {$\pm$ 1.7} & 20.0 {$\pm$ 5.1} & 40.1 {$\pm$ 1.8} & 29.5 {$\pm$ 6.1} & 40.1 {$\pm$ 1.8} & 31.6 {$\pm$ 0.8} \\

        & WaveGNN & {\underline{43.2 $\pm$ { 3.5}}} & \textbf{{46.6 $\pm$ { 6.1}}} & \underline{{43.2 $\pm$ { 3.5}}} & \textbf{{38.2 $\pm$ { 3.4}}} & \underline{51.2$\pm$ { 3.3}} & \textbf{63.3 $\pm$ { 5.1}} & \underline{51.2 $\pm$ { 3.3}} & \textbf{51.2 $\pm$ { 3.3}} \\
        \cmidrule{2-10}
        & \textit{Improvement} & -7.29\% & +4.71\% & -0.91\% & +0.52\% & -4.47\% & +6.56\% & -4.47\% & +0.99\% \\
        \bottomrule
    \end{tabular}%
    }
    \vspace{-5pt}
\end{table*}

\subsection{Results} \label{sec:results}

\textbf{Overall Performance.} We report the overall performance of WaveGNN and competing baselines in Tables~\ref{tab:overall_results_1} and~\ref{tab:mimic_results}, highlighting the \textbf{best} and \underline{second best} scores.

WaveGNN achieves the best performance on PAM across all metrics, underscoring its strength in multi-class activity classification under high sparsity. On P12, it provides the top AUPRC (49.4) and an AUROC nearly tied with DGM$^{2}$-O (83.9 vs.\ 84.4, a 0.5 gap), showing competitive performance on highly irregular mortality prediction. On P19, GraFITi achieves higher scores, because this binary task involves much shorter sequences, favoring interpolation-based methods.

As shown in Table~\ref{tab:mimic_results}, on the MIMIC dataset, WaveGNN shows mixed results on the binary 48-IHM task. While WaveGNN improves over the next best model, UTDE, by 2.64\% in F1 score, UTDE achieves a higher AUPRC. This can be attributed to MIMIC's lower missingness compared to P12 and P19, which makes interpolation-based models more competitive in this relatively easier binary task. However, in the more complex multi-label classification task of 24-PHE on the same dataset, WaveGNN outperforms UTDE by \emph{23.7\%} in F1 score, suggesting that imputation-based approaches are less effective in challenging, multi-label settings.

Among graph-based baselines, Raindrop and GraFITi achieve competitive results in specific tasks (e.g., GraFITi on P19), but their performance fluctuates sharply across datasets. Conversely, WaveGNN consistently ranks among the top methods under varying levels of irregularity and sequence lengths.

\textbf{Consistency Across Datasets and Tasks.} To evaluate consistency in performance beyond individual benchmarks, we compute the average rank of each method across all 12 dataset-metric pairs. Table~\ref{tab:overall-res-summary} shows the results. WaveGNN achieves the lowest average rank (1.67), substantially ahead of the next best baseline (UTDE at 4.58). Moreover, WaveGNN is ranked first in 7 cases and within the top two in 9 cases, whereas competing methods rarely maintain such consistency (e.g., UTDE achieves top performance on some MIMIC evaluation metrics but ranks poorly on P12 and PAM, GraFITi excels on P19 but drops sharply on P12). These results demonstrate that WaveGNN provides 
\emph{consistent performance across datasets and tasks}, avoiding the inconsistency 
of baselines that perform well only under favorable conditions.

\textbf{Robustness to Missing Data.}\label{exp:leave-sensors} 
We evaluate WaveGNN under more challenging conditions where entire sensor streams are missing, simulating real-world scenarios such as sensor failures or deployment constraints. Following Raindrop~\cite{zhang2021graph}, we experiment with two setups on the PAM dataset: (1) removing all observations from a fixed set of the most informative sensors, and (2) removing observations from a randomly selected set of sensors. Table~\ref{tab:pam_results_2_3} reports the results for both settings when 10\%, 30\%, and 50\% of sensors are removed.

WaveGNN outperforms all baselines in 20 out of 24 experiments and ranks second in the remaining four. Notably, it achieves up to 5\% relative improvement in F1 score with 10\% sensor removal, and gains of 12.3\% and 6.5\% with 30\% and 50\% removal, respectively.

Among the baselines, \textit{Raindrop} and \textit{WarpFormer} are the most competitive in robustness. In contrast, \textit{GraFITi}, which previously ranked among the top performers, suffers a significant performance drop. This drop is likely due to GraFITi’s reliance on inter-variable interactions at shared timestamps. When informative sensor observations are missing, the model has less reliable information to compensate for irregularity.

Imputation-based methods such as \textit{mTAND} and \textit{UTDE} also experience substantial declines in robustness with increasing sensor removal. These approaches still try to fill in the missing values, but with entire sensor observations missing, they have little to no reliable data to base their imputations on. This leads to inaccurate or biased reconstructions which affects their robustness under missing data.

\subsection{Ablation Study}
We created different variations of WaveGNN to understand the effectiveness of its components: \textit{w/o short-term} and \textit{w/o long-term} variants remove the $A_S$ and $A_L$ similarity matrices, so that the short- and long-term inter-series relationships are not considered, respectively. The \textit{w/o inter-series} variant excludes the graph message passing part, so neither of the inter-series similarity measures are considered, while the \textit{w/o intra-series} variant removes the transformer-based module and simply averages the observation embeddings for each sensor to obtain a univariate sequence embedding. Finally, the \textit{w/o temporal encoding} variant discards the temporal encoding module and uses a simple positional encoding for the Transformer, and the \textit{w/o decay rate} variant removes WaveGNN's decay mechanism, aggregating the output sequence embedding of the Transformer with equal weights, thereby treating all sequence observations equally.

Table~\ref{tab:ablation_pam_results} presents the results of the ablation study on the PAM dataset, showing the impact of removing various components of WaveGNN. Our findings indicate that all components contribute to improved performance, with the most significant drops occurring when inter-series and intra-series relationships are removed, leading to performance decreases of 35.3\% and 15.1\% in weighted F1 score, respectively. This shows the importance of simultaneously addressing both intra-variable dynamics and inter-variable dependencies to effectively model irregular datasets.

Interestingly, the removal of dynamic short-term dependencies has a greater negative impact on performance than the removal of long-term dependencies. This outcome is intuitive, as short-term dependencies between sensors within the current input window are often more critical than long-term dependencies, given their specificity to the individual sample and time window, rather than being generalized across all samples and time periods. For example, the relationship between a patient's heart rate and blood sugar level during a particular time window might differ from the typical pattern usually seen in that patient, possibly due to the effects of medication or other factors, making it more relevant to focus on these short-term dependencies.

Lastly, both the temporal encoding and the decay mechanism components are integral to WaveGNN's overall performance. Temporal encoding allows the model to capture the temporal dynamics and patterns within each sensor’s time series, while the decay mechanism ensures that more recent data, which is often more indicative of final outcomes, is given appropriate weight. Together, these components enable WaveGNN to manage the timing and relevance of observations effectively, contributing to overall performance.

\begin{figure*}[t!]
    \centering
    \subfloat[Dynamic Short-Term Dependencies ($A_S$) - Sample 1]{%
        \includegraphics[width=0.32\textwidth]{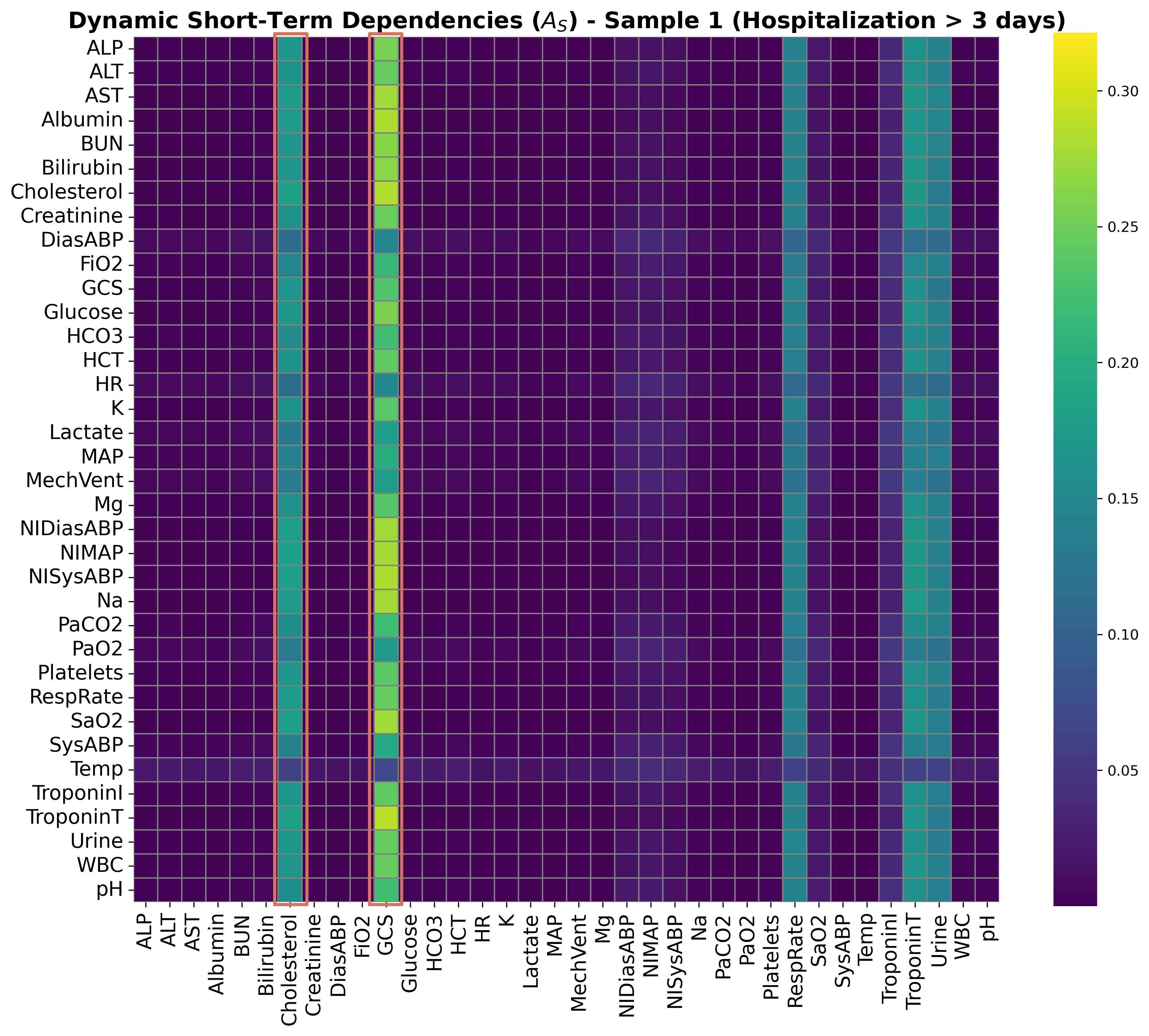}%
        \label{fig:adj_dyn1}%
    }\hfill
    \subfloat[Dynamic Short-Term Dependencies ($A_S$) - Sample 2]{%
        \includegraphics[width=0.32\textwidth]{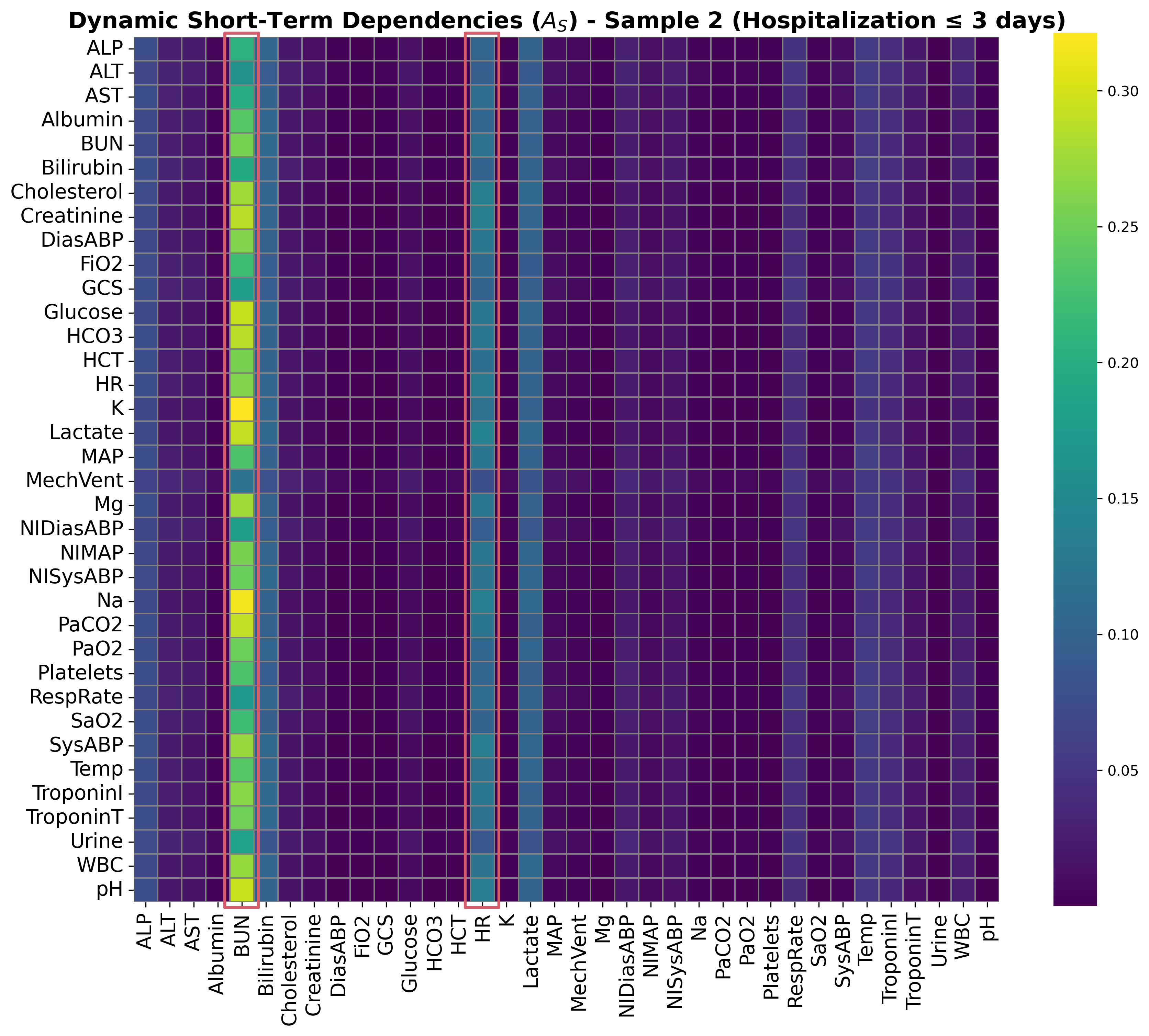}%
        \label{fig:adj_dyn2}%
    }
    \subfloat[Static Long-Term Similarities ($A_L$)]{%
        \includegraphics[width=0.32\textwidth]{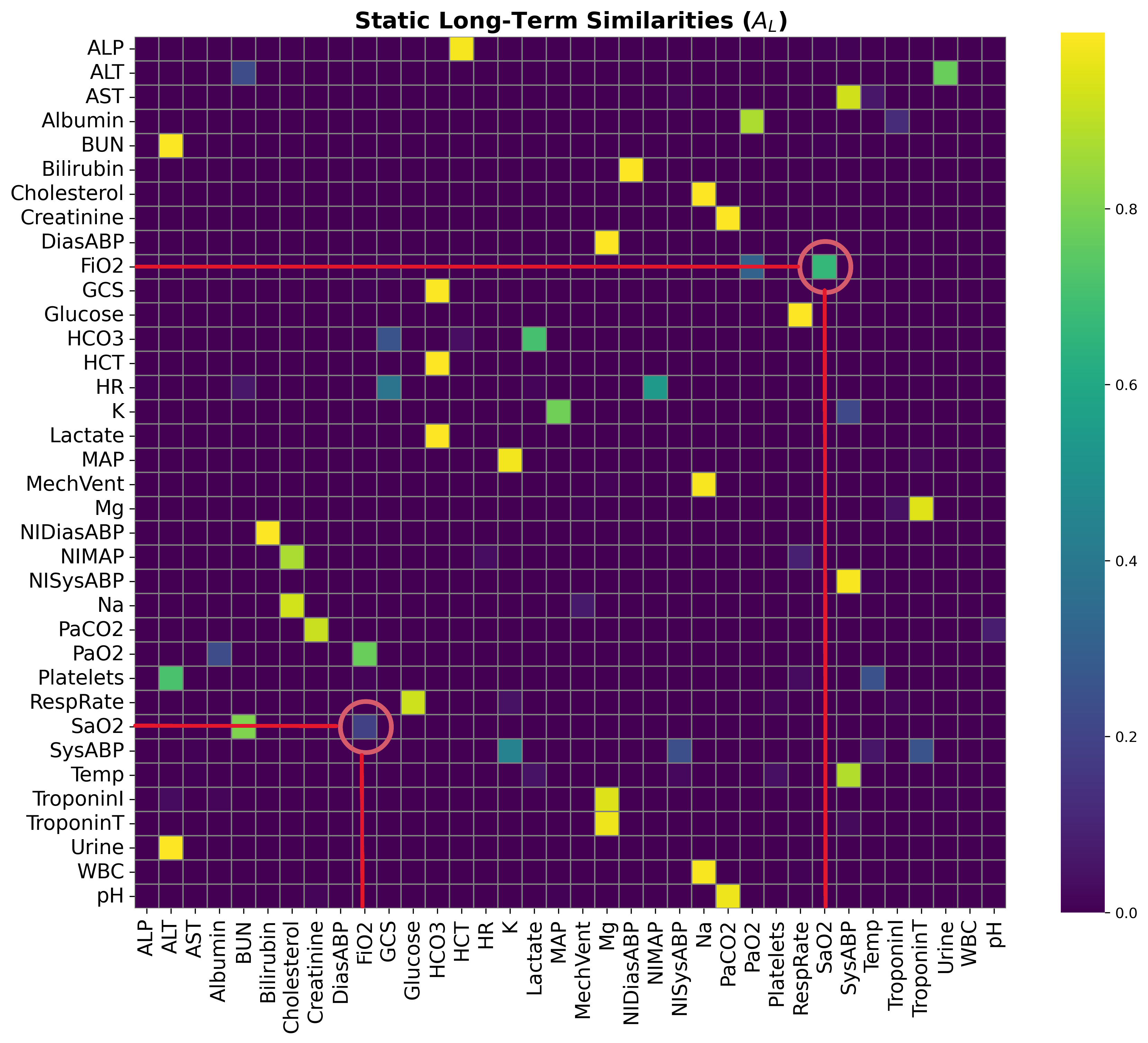}%
        \label{fig:adj_static}%
    }
    \caption{Learned adjacency matrices capturing feature dependencies in ICU patients. 
(a) and (b) show dynamic short-term dependencies ($A_S$) for two test samples, highlighting patient-specific variations. 
(c) presents the static long-term similarity matrix ($A_L$), representing global feature relationships shared across all samples.}
    \label{fig:adj_mat}
    \vspace{-13pt}
\end{figure*}

\begin{table}[ht]
    \centering
    \caption{Ablation study results on the PAM dataset. We report the
average results across 3 independent runs.}
    \resizebox{0.5\textwidth}{!}{%
    \label{tab:ablation_pam_results}
    \begin{tabular}{c|c c c c}
        \toprule
        {\textbf{Ablation Setting}} & \textbf{Accuracy} & \textbf{Precision} & \textbf{Recall} & \textbf{F1 score} \\
        \midrule
        w/o short-term & {89.9 { $\pm$ 1.9}} & {90.1 { $\pm$ 1.9}} & {89.9 { $\pm$ 1.9}} &  {89.8 { $\pm$ 2.0}} \\
         w/o long-term & {92.3 { $\pm$ 1.9}} & {92.4 { $\pm$ 1.9}} & {92.3 { $\pm$ 1.9}} &  {92.2 { $\pm$ 1.9}} \\
          w/o inter-series & {64.1 { $\pm$ 1.7}} & {63.5 { $\pm$ 2.0}} & {64.1 { $\pm$ 3.1}} &  {61.9 { $\pm$ 3.1}} \\
          w/o intra-series & {81.2 { $\pm$ 1.3}} & {81.5 { $\pm$ 1.5}} & {81.2 { $\pm$ 1.3}} &  {81.2 { $\pm$ 1.4}} \\
          w/o temporal encoding & {90.6 {  $\pm$ 1.1}} & {90.6 { $\pm$ 1.0}} & {90.6 { $\pm$ 1.1}} &  {90.4 { $\pm$ 1.2}} \\
          w/o decay rate & {90.6 {  $\pm$ 2.3}} & {90.7 { $\pm$ 2.3}} & {90.6 { $\pm$ 2.3}} &  {90.5 { $\pm$ 2.3}} \\
          \midrule
          WaveGNN & \textbf{95.6 {$\pm$ 1.1}} & \textbf{95.7 {$\pm$ 1.1}} & \textbf{95.6 {$\pm$ 1.1}} & \textbf{95.6 {$\pm$ 1.1}} \\
        \bottomrule
    \end{tabular}%
    }
    \vspace{-10 pt}
\end{table}

\subsection{Interpretation of Learned Sensor Graphs}\label{sec:interpretability}

In this section, we examine the interpretability of the sensor graphs constructed by WaveGNN, using a trained model checkpoint on the P12 dataset. Specifically, we analyze the adjacency matrices learned for two distinct test samples, which were excluded from the training set. The first sample corresponds to a patient who remained in the ICU for more than three days (\( \text{label} = 1 \)), while the second corresponds to a patient discharged within three days (\( \text{label} = 0 \)).

As detailed in Section~\ref{sec:meth}, WaveGNN constructs two types of adjacency matrices to model graph-based relationships: a static, globally shared matrix \( A_L \), encoding long-term similarities, and dynamic, sample-specific matrix \( A_S \), encoding short-term dependencies within the input window. These matrices capture directed pairwise relationships between sensor features, where \( A_{i,j} \) represents the influence of feature \( i \) on feature \( j \). The final adjacency matrix is derived as a weighted combination of \( A_L \) and \( A_S \), controlled by a trainable parameter \( \alpha \) (Equation~\ref{eq:adj}).

Figure~\ref{fig:adj_mat} presents heatmaps of the learned adjacency matrices. Rows and columns represent sensor features, while the color intensity depicts the edge weight, indicating the strength of pairwise feature relationships. 

\paragraph{Static Long-Term Similarities (\( A_L \))}  
The globally shared matrix \( A_L \) (Figure~\ref{fig:adj_static}) captures consistent long-term feature dependencies across the dataset. For example, FiO\textsubscript{2} (fraction of inspired oxygen) and SaO\textsubscript{2} (oxygen saturation) exhibit a relatively strong relationship (highlighted with red circles), aligned with the clinical understanding that FiO\textsubscript{2} directly impacts blood oxygen levels~\cite{severinghaus1979simple}. This global dependency suggests that WaveGNN effectively captures universal feature relationships important for modeling patient states.

\paragraph{Dynamic Short-Term Dependencies (\( A_S \))}  
The dynamic matrices \( A_S \) (Figures~\ref{fig:adj_dyn1} and~\ref{fig:adj_dyn2}) are generated for each sample by WaveGNN’s learned graph constructor, adjusting the graph to each patient's unique characteristics. These matrices encode short-term dependencies that vary across patients:
\begin{itemize}
    \item For the sample with \( \text{label} = 1 \) (Figure~\ref{fig:adj_dyn1}), Glasgow Coma Scale (GCS) and Cholesterol exhibit strong connections to other features (highlighted with red bars), suggesting their importance in predicting prolonged ICU stays. GCS is widely used to assess neurological status, and lower scores are often associated with critical conditions requiring extended ICU care~\cite{teasdale1974assessment}. Cholesterol levels are linked to metabolic and inflammatory responses, which can influence recovery time, with abnormalities often observed in patients with prolonged hospitalization~\cite{linton2019role}. The model’s identification of these relationships aligns with clinical expectations, indicating that WaveGNN effectively captures key physiological dependencies associated with extended ICU stays.
    \item For the sample with \( \text{label}=0 \) (Figure~\ref{fig:adj_dyn2}), Blood Urea Nitrogen (BUN) and Heart Rate (HR) exhibit strong connections to other features (highlighted with red bars), suggesting their role in characterizing shorter ICU stays. BUN is commonly used to monitor kidney function and metabolic stress, while HR is a key vital sign reflecting physiological stability. The learned graph structure indicates that these features are highly interconnected in short-stay patients, aligning with the idea that acute physiological changes tend to be more immediate and short-lived~\cite{hosten1990bun}. In contrast, prolonged ICU stays often involve more complex dependencies beyond acute variations, as seen in the previous case with GCS and Cholesterol. 
    These differences in feature interactions for the two samples suggest that WaveGNN effectively captures short-term physiological variations, adjusting its graph structure to reflect patient-specific ICU trajectories.
\end{itemize}

Notably, the values in the short-term dependency matrices (\( A_S \)) are smaller compared to the long-term similarity matrix (\( A_L \)), as \( A_S \) refines the graph structure for each sample rather than encoding global dependencies. The balance between these components is controlled by the learned parameter \( \alpha \), which determines the relative contributions of \( A_L \) and \( A_S \) in the final graph. For this trained checkpoint, the model has learned \( \alpha = 0.5211 \), indicating that short-term dependencies (\( A_S \)) contribute slightly more than long-term similarities (\( A_L \)).

\section{Conclusion}
\label{sec:conclusion}
We introduced WaveGNN, a novel framework for modeling irregular multivariate time series that avoids imputation and operates directly on raw clinical data. By combining a decay-aware Transformer for intra-series encoding with a sample-specific graph neural network for inter-series modeling, WaveGNN produces sparse, interpretable graphs per instance. Our evaluation shows that WaveGNN achieves \emph{consistent performance} across four clinical benchmarks, obtaining the lowest average rank across 12 evaluations. The model is also \emph{robust} to varying levels of missingness, outperforming all baselines in 20 of 24 missing-data robustness experiments, while providing interpretable graph structures that align with known physiological relationships. In the future, we plan to expand WaveGNN to support multimodal scenarios.

\bibliographystyle{IEEEtran}
\bibliography{wavegnn}

\begin{thebibliography}{10}
\providecommand{\url}[1]{#1}
\csname url@samestyle\endcsname
\providecommand{\newblock}{\relax}
\providecommand{\bibinfo}[2]{#2}
\providecommand{\BIBentrySTDinterwordspacing}{\spaceskip=0pt\relax}
\providecommand{\BIBentryALTinterwordstretchfactor}{4}
\providecommand{\BIBentryALTinterwordspacing}{\spaceskip=\fontdimen2\font plus
\BIBentryALTinterwordstretchfactor\fontdimen3\font minus \fontdimen4\font\relax}
\providecommand{\BIBforeignlanguage}[2]{{%
\expandafter\ifx\csname l@#1\endcsname\relax
\typeout{** WARNING: IEEEtran.bst: No hyphenation pattern has been}%
\typeout{** loaded for the language `#1'. Using the pattern for}%
\typeout{** the default language instead.}%
\else
\language=\csname l@#1\endcsname
\fi
#2}}
\providecommand{\BIBdecl}{\relax}
\BIBdecl

\bibitem{zhang2021graph}
X.~Zhang, M.~Zeman, T.~Tsiligkaridis, and M.~Zitnik, ``Graph-guided network for irregularly sampled multivariate time series,'' in \emph{International Conference on Learning Representations}, 2021.

\bibitem{horn2020set}
M.~Horn, M.~Moor, C.~Bock, B.~Rieck, and K.~Borgwardt, ``Set functions for time series,'' in \emph{International Conference on Machine Learning}.\hskip 1em plus 0.5em minus 0.4em\relax PMLR, 2020, pp. 4353--4363.

\bibitem{lipton2016directly}
Z.~C. Lipton, D.~Kale, and R.~Wetzel, ``Directly modeling missing data in sequences with rnns: Improved classification of clinical time series,'' in \emph{Machine learning for healthcare conference}.\hskip 1em plus 0.5em minus 0.4em\relax PMLR, 2016, pp. 253--270.

\bibitem{shukla2018interpolation}
S.~N. Shukla and B.~M. Marlin, ``Interpolation-prediction networks for irregularly sampled time series,'' \emph{arXiv preprint arXiv:1909.07782}, 2019.

\bibitem{shukla2020multi}
S.~N. Shukla and B.~Marlin, ``Multi-time attention networks for irregularly sampled time series,'' in \emph{International Conference on Learning Representations}, 2020.

\bibitem{che2018recurrent}
Z.~Che, S.~Purushotham, K.~Cho, D.~Sontag, and Y.~Liu, ``Recurrent neural networks for multivariate time series with missing values,'' \emph{Scientific reports}, vol.~8, no.~1, p. 6085, 2018.

\bibitem{rubanova2019latent}
Y.~Rubanova, R.~T. Chen, and D.~K. Duvenaud, ``Latent ordinary differential equations for irregularly-sampled time series,'' \emph{Advances in neural information processing systems}, vol.~32, 2019.

\bibitem{mcdermott2021comprehensive}
M.~McDermott, B.~Nestor, E.~Kim, W.~Zhang, A.~Goldenberg, P.~Szolovits, and M.~Ghassemi, ``A comprehensive ehr timeseries pre-training benchmark,'' in \emph{Proceedings of the Conference on Health, Inference, and Learning}, 2021, pp. 257--278.

\bibitem{zhang2024irregular}
W.~Zhang, C.~Yin, H.~Liu, X.~Zhou, and H.~Xiong, ``Irregular multivariate time series forecasting: A transformable patching graph neural networks approach,'' in \emph{Forty-first International Conference on Machine Learning}, 2024.

\bibitem{zhang2023warpformer}
J.~Zhang, S.~Zheng, W.~Cao, J.~Bian, and J.~Li, ``Warpformer: A multi-scale modeling approach for irregular clinical time series,'' in \emph{Proceedings of the 29th ACM SIGKDD Conference on Knowledge Discovery and Data Mining}, 2023, pp. 3273--3285.

\bibitem{yalavarthi2024grafiti}
V.~K. Yalavarthi, K.~Madhusudhanan, R.~Scholz, N.~Ahmed, J.~Burchert, S.~Jawed, S.~Born, and L.~Schmidt-Thieme, ``Grafiti: Graphs for forecasting irregularly sampled time series,'' in \emph{Proceedings of the AAAI Conference on Artificial Intelligence}, vol.~38, no.~15, 2024, pp. 16\,255--16\,263.

\bibitem{wu2021dynamic}
Y.~Wu, J.~Ni, W.~Cheng, B.~Zong, D.~Song, Z.~Chen, Y.~Liu, X.~Zhang, H.~Chen, and S.~B. Davidson, ``Dynamic gaussian mixture based deep generative model for robust forecasting on sparse multivariate time series,'' in \emph{Proceedings of the AAAI Conference on Artificial Intelligence}, vol.~35, no.~1, 2021, pp. 651--659.

\bibitem{zhang2023improving}
X.~Zhang, S.~Li, Z.~Chen, X.~Yan, and L.~R. Petzold, ``Improving medical predictions by irregular multimodal electronic health records modeling,'' in \emph{International Conference on Machine Learning}.\hskip 1em plus 0.5em minus 0.4em\relax PMLR, 2023, pp. 41\,300--41\,313.

\bibitem{hajisafi2023learning}
A.~Hajisafi, H.~Lin, S.~Shaham, H.~Hu, M.~D. Siampou, Y.-Y. Chiang, and C.~Shahabi, ``Learning dynamic graphs from all contextual information for accurate point-of-interest visit forecasting,'' in \emph{Proceedings of the 31st ACM International Conference on Advances in Geographic Information Systems}, 2023, pp. 1--12.

\bibitem{hajisafi2024dynamic}
A.~Hajisafi, H.~Lin, Y.-Y. Chiang, and C.~Shahabi, ``Dynamic gnns for precise seizure detection and classification from eeg data,'' in \emph{Pacific-Asia Conference on Knowledge Discovery and Data Mining}.\hskip 1em plus 0.5em minus 0.4em\relax Springer, 2024, pp. 207--220.

\bibitem{velivckovic2018graph}
P.~Veli{\v{c}}kovi{\'c}, G.~Cucurull, A.~Casanova, A.~Romero, P.~Li{\`o}, and Y.~Bengio, ``Graph attention networks,'' in \emph{International Conference on Learning Representations}, 2018.

\bibitem{kazemi2019time2vec}
S.~M. Kazemi, R.~Goel, S.~Eghbali, J.~Ramanan, J.~Sahota, S.~Thakur, S.~Wu, C.~Smyth, P.~Poupart, and M.~Brubaker, ``Time2vec: Learning a vector representation of time,'' \emph{arXiv preprint arXiv:1907.05321}, 2019.

\bibitem{vaswani2017attention}
A.~Vaswani, N.~Shazeer, N.~Parmar, J.~Uszkoreit, L.~Jones, A.~N. Gomez, {\L}.~Kaiser, and I.~Polosukhin, ``Attention is all you need,'' \emph{Advances in neural information processing systems}, vol.~30, 2017.

\bibitem{velivckovic2017graph}
P.~Veli{\v{c}}kovi{\'c}, G.~Cucurull, A.~Casanova, A.~Romero, P.~Lio, and Y.~Bengio, ``Graph attention networks,'' \emph{arXiv preprint arXiv:1710.10903}, 2017.

\bibitem{kipf2017semisupervised}
T.~N. Kipf and M.~Welling, ``Semi-supervised classification with graph convolutional networks,'' in \emph{ICLR}, 2017.

\bibitem{chen2020measuring}
D.~Chen and et~al., ``Measuring and relieving the over-smoothing problem for graph neural networks from the topological view,'' in \emph{AAAI}, vol.~34, no.~04, 2020, pp. 3438--3445.

\bibitem{goldberger2000physiobank}
A.~L. Goldberger, L.~A. Amaral, L.~Glass, J.~M. Hausdorff, P.~C. Ivanov, R.~G. Mark, J.~E. Mietus, G.~B. Moody, C.-K. Peng, and H.~E. Stanley, ``Physiobank, physiotoolkit, and physionet: components of a new research resource for complex physiologic signals,'' \emph{circulation}, vol. 101, no.~23, pp. e215--e220, 2000.

\bibitem{johnson2016mimic}
A.~E. Johnson, T.~J. Pollard, L.~Shen, L.-w.~H. Lehman, M.~Feng, M.~Ghassemi, B.~Moody, P.~Szolovits, L.~Anthony~Celi, and R.~G. Mark, ``Mimic-iii, a freely accessible critical care database,'' \emph{Scientific data}, vol.~3, no.~1, pp. 1--9, 2016.

\bibitem{reyna2020early}
M.~A. Reyna, C.~S. Josef, R.~Jeter, S.~P. Shashikumar, M.~B. Westover, S.~Nemati, G.~D. Clifford, and A.~Sharma, ``Early prediction of sepsis from clinical data: the physionet/computing in cardiology challenge 2019,'' \emph{Critical care medicine}, vol.~48, no.~2, pp. 210--217, 2020.

\bibitem{reiss2012introducing}
A.~Reiss and D.~Stricker, ``Introducing a new benchmarked dataset for activity monitoring,'' in \emph{2012 16th international symposium on wearable computers}.\hskip 1em plus 0.5em minus 0.4em\relax IEEE, 2012, pp. 108--109.

\bibitem{harutyunyan2019multitask}
H.~Harutyunyan, H.~Khachatrian, D.~C. Kale, G.~Ver~Steeg, and A.~Galstyan, ``Multitask learning and benchmarking with clinical time series data,'' \emph{Scientific data}, vol.~6, no.~1, p.~96, 2019.

\bibitem{wu2020connecting}
Z.~Wu, S.~Pan, G.~Long, J.~Jiang, X.~Chang, and C.~Zhang, ``Connecting the dots: Multivariate time series forecasting with graph neural networks,'' in \emph{Proceedings of the 26th ACM SIGKDD international conference on knowledge discovery \& data mining}, 2020, pp. 753--763.

\bibitem{loshchilov2017decoupled}
I.~Loshchilov and F.~Hutter, ``Decoupled weight decay regularization,'' \emph{arXiv preprint arXiv:1711.05101}, 2017.

\bibitem{severinghaus1979simple}
J.~W. Severinghaus, ``Simple, accurate equations for human blood o2 dissociation computations,'' \emph{Journal of Applied Physiology}, vol.~46, no.~3, pp. 599--602, 1979.

\bibitem{teasdale1974assessment}
G.~Teasdale and B.~Jennett, ``Assessment of coma and impaired consciousness: a practical scale,'' \emph{The Lancet}, vol. 304, no. 7872, pp. 81--84, 1974.

\bibitem{linton2019role}
M.~F. Linton, P.~G. Yancey, S.~S. Davies, W.~G. Jerome, E.~F. Linton, W.~L. Song, A.~C. Doran, and K.~C. Vickers, ``The role of lipids and lipoproteins in atherosclerosis,'' \emph{Endotext [Internet]}, 2019.

\bibitem{hosten1990bun}
A.~O. Hosten, ``Bun and creatinine,'' \emph{Clinical Methods: The History, Physical, and Laboratory Examinations. 3rd edition}, 1990.

\end{thebibliography}

\end{document}